\documentclass[runningheads]{llncs}

\usepackage{eccv}

\usepackage{eccvabbrv}

\usepackage{graphicx}
\usepackage{booktabs}

\usepackage[accsupp]{axessibility}  % Improves PDF readability for those with disabilities.

\usepackage{graphicx}
\usepackage{multirow}
\usepackage{wrapfig}
\usepackage{amsmath}
\usepackage{caption}
\usepackage{subcaption}
\usepackage{enumitem}
\usepackage{graphicx}
\usepackage{caption}
\usepackage{color}
\usepackage[most]{tcolorbox}

\usepackage[dvipsnames]{xcolor}
\usepackage{colortbl}
\usepackage{color}

\usepackage{arydshln}

\usepackage{adjustbox}
\usepackage[misc]{ifsym}

\usepackage[dvipsnames]{xcolor}
\usepackage[pagebackref,breaklinks,colorlinks,citecolor=eccvblue]{hyperref}
\usepackage{orcidlink}

\begin{document}

\title{Multi-modal Generation via Cross-Modal In-Context Learning} 


\author{Amandeep Kumar\textsuperscript{1}  \quad Muzammal Naseer\textsuperscript{1} \quad Sanath Narayan\textsuperscript{2} \\ \quad  Rao Muhammad Anwer\textsuperscript{1} \quad Salman Khan\textsuperscript{1} Hisham Cholakkal\textsuperscript{1} 
}

\authorrunning{Kumar et al.}

\institute{\textsuperscript{1}Mohamed bin Zayed University of AI \quad \textsuperscript{2}Technology Innovation Institute
\email{amandeep.kumar@mbzuai.ac.ae}} 

\maketitle

\begin{center}
        \centering
        \captionsetup{type=figure}
        {\includegraphics[width=1.0\textwidth]{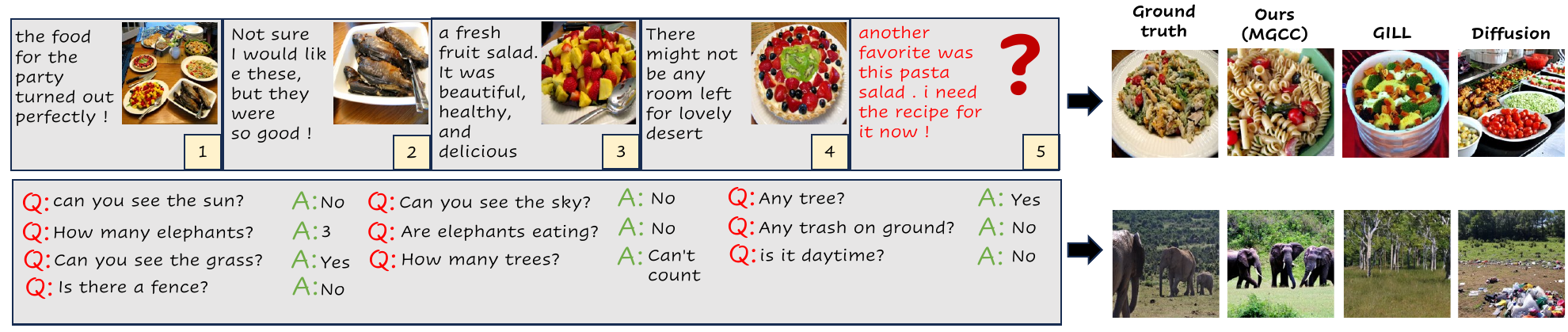}}\vspace{-0.0cm}
        \caption{Example illustration of image generation based on complex multimodal prompt sequences. The top row shows challenges faced by diffusion in aligning the image with the prompt, while state-of-the-art GILL generates a holistic image combining all prompts. Our method surpasses both by aligning with the final text and incorporating the visual appearance of \textit{pasta"} from the first image in the sequence, resulting in a more contextually accurate image. In the bottom row, both diffusion and GILL fail to capture the context of the mentioned object in the dialogue. In contrast, our method generates the \textit{elephant"} as described in the dialogue and preserves count information, illustrating a comprehensive and context-aware image generation process (best viewed in zoom). }
        \label{intro-image}
    \end{center}

\begin{abstract}
In this work, we study the problem of generating novel images from complex multimodal prompt sequences. While existing methods achieve promising results for text-to-image generation, they often struggle to capture fine-grained details from lengthy prompts and maintain contextual coherence within prompt sequences. Moreover, they often result in misaligned image generation for prompt sequences featuring multiple objects. To address this, we propose a Multi-modal Generation via Cross-Modal In-Context Learning (MGCC) method that generates novel images from complex multimodal prompt sequences by leveraging the combined capabilities of large language models (LLMs) and diffusion models. Our MGCC comprises a novel Cross-Modal Refinement module to explicitly learn cross-modal dependencies between the text and image in the LLM embedding space, and a contextual object grounding module to generate object bounding boxes specifically targeting  scenes with multiple objects. Our MGCC demonstrates a diverse range of multimodal capabilities, like novel image generation, the facilitation of multimodal dialogue, and generation of texts. Experimental evaluations on two benchmark datasets,  demonstrate the effectiveness of our method. On Visual Story Generation (VIST) dataset with multimodal inputs, our MGCC achieves a CLIP Similarity score of $0.652$ compared to SOTA  GILL $0.641$. Similarly, on Visual Dialogue Context (VisDial) having lengthy dialogue sequences, our MGCC achieves an impressive CLIP score of $0.660$, largely outperforming existing SOTA method scoring $0.645$. Code: \url{https://github.com/VIROBO-15/MGCC}
\end{abstract}

\section{Introduction}
\label{sec:intro}

The advancement of large language models (LLMs) \cite{openai2023gpt, liu2023visual} trained on extensive textual corpora has enabled remarkable adaptability across various modalities. Earlier works demonstrated the effectiveness of grounding text-only LLMs to images for vision-and-language tasks \cite{zhu2023minigpt, chen2023minigptv2, tsimpoukelli2021multimodal, ilharco2020probing, li2023blip}, as well as in embodied settings for robotics \cite{driess2023palm, ahn2022can} and beyond. These methods leverage the capabilities of LLMs that are trained on large scale text-only data, while keeping the LLM weights frozen. 
In this work, we tackle the problem of generating novel images with lengthy text descriptions or complex sequence of text prompts by leveraging the capabilities of both LLMs \cite{zhang2022opt} and diffusion models \cite{ramesh2022hierarchical, rombach2022high}. 

% Our proposed approach, \textbf{M}ulti-modal \textbf{G}eneration via \textbf{C}ross-Modal In-\textbf{C}ontext Learning(MGCC), is capable of generating multimodal outputs (texts and images). 
Recent advances in text-to-image generation methods \cite{rombach2022high, ramesh2022hierarchical, saharia2022photorealistic, shamshad2023clip2protect} have demonstrated 
impressive results in generating novel images. However, these approaches tend to overlook fine-grained details in the case of lengthy text prompts or complex text sequences that require understanding the previous context of the prompts.

Generally, these approaches struggle in these scenarios likely due to two reasons: (a) they rely on the CLIP text encoder \cite{radford2021learning} that is limited to handling 77 tokens at a time, leading to loss of crucial information in lengthy text prompts, and (b) they cannot process interleaved text-image sequences as input. To overcome these challenges, GILL \cite{koh2023generating} proposed to utilize pretrained LLMs. 
% To take images as the input, first it learns to translate input images in the LLM voculabary space. 
First, for handling image inputs, it learns to transform images to the LLM vocabulary space. 
Then, to generate images, it aligns the LLM output embedding space to CLIP text encoder output space \cite{radford2021learning} via a transformer encoder-decoder module \cite{vaswani2017attention}. Such an alignment allows conditioning diffusion on the LLM embedding for generating images.

While GILL is capable of generating images with lengthy prompt descriptions and complex sequence of prompts, it still struggles to generate accurate images aligned with the prompts sequence. This can be attributed to the use of pre-trained LLMs that are implicitly designed to handle dependence within the sequence token but not explicitly designed for handling the cross-modal context, such as image and text tokens. A straightforward solution is to fine-tune the LLM.  However, fine-tuning the LLM requires large amount of interleaved image text pairs and extensive compute resource \cite{alayrac2022flamingo, aghajanyan2022cm3}. This approach can also lead to a loss of generalization, which was learned from the large text corpus. In this work, we address the aforementioned limitations by training on the image-captions \cite{aghajanyan2022cm3} alone.

\noindent \textbf{Contributions:} We propose an approach named \textbf{M}ulti-modal \textbf{G}eneration via \textbf{C}ross-Modal In-\textbf{C}ontext Learning (MGCC) that learns to generate multimodal outputs given lengthy multimodal inputs. To this end, we introduce a novel \textit{Cross-Modal Refinement Module} to enable learning the cross-modal dependencies between text and image in the LLMs embedding space during training. This module aids the pre-trained LLM to explicitly learn the correspondence between text and image tokens using cross-attentions. By leveraging the refinement module, the model gains semantic understanding of the scene based on the input prompt sequence. Moreover, to enhance the fine grained details in the output, we incorporate a  \textit{contextual object grounding module}. Utilizing the in-context learning \cite{lian2023llm, brooks2023instructpix2pix}, we predict bounding boxes of the objects present in the prompt while maintaining the temporal consistency of the prompt sequence. Thereby, we collectively solve the problem of the object present in the scene and their count.% the associated object count.

Extensive quantitative and qualitative experiments are conducted on two datasets: Visual Story Generation (VIST) \cite{huang2016visual} and Visual Dialogue Context (VisDial) \cite{das2017visual}. Our MGCC performs favorably against text-to-image generation methods and state-of-the-art  GILL \cite{koh2023generating}. When handling multimodal context in VIST dataset, our MGCC outperforms the state-of-the-art approach in terms of both CLIP Similarity from $0.641$ to $0.652$ and LPIPS score from $0.693$ to $0.679$. Similarly, on challenging VisDial dataset with long dialogue prompts, our MGCC achieves a CLIP Similarity score of $0.660$  largely outperforming the SOTA GILL  with $0.645$. Fig. \ref{intro-image} shows the generation of novel images by our MGCC, illustrating the improved alignment of our generated images with the prompts while maintaining temporal consistency.  

\section{Related Works:}

\begin{figure*}[t!]
\centering
\includegraphics[width=1\textwidth]{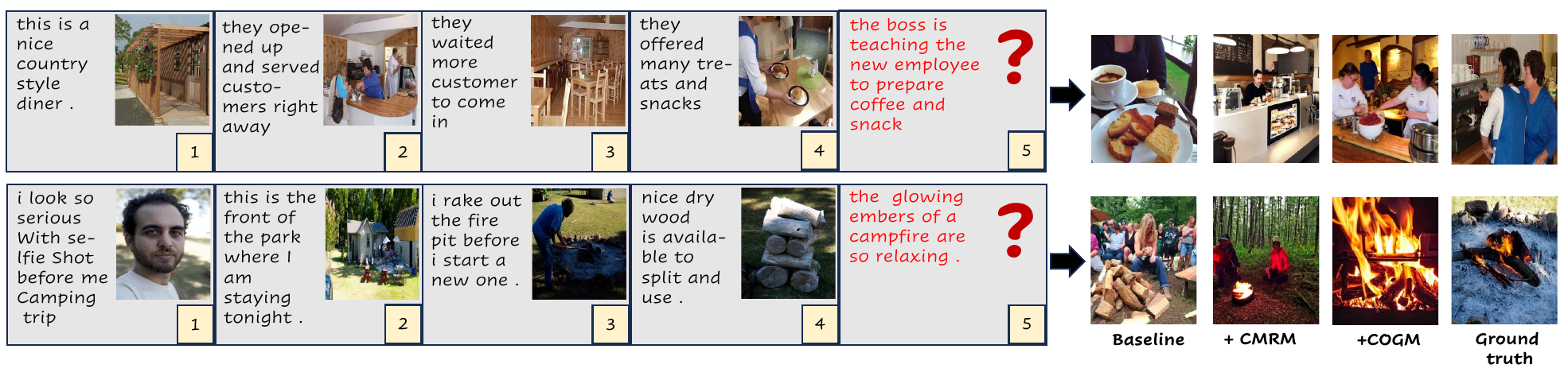}
\caption{Example images depicting the impact of progressively integrating our cross-modal refinement module  (CMRM) and contextual object grounding module (COGM)   into the baseline. In  \textbf{\textit{first row}}, the baseline generates an image of  \textit{``cookies and coffee in a plate"} which doesn't align with the earlier prompts \textit{``the boss is teaching the new employee to prepare coffee and snack."} Although the integration of our CMRM module to baseline improves semantic understanding, the generated image still fails to include the person instance in the scene. Finally,  by incorporating our GCO (grounding contextual objects), we achieve better alignment with the ground truth, resulting in an image that accurately matches the number of \textit{``persons"} mentioned in the earlier prompt. Similarly, in \textbf{\textit{second row}}, baseline struggles to generate an image consistent with the text \textit{``the glowing embers of a campfire is so relaxing"}. Our refinement module comprehends the prompts and generates \textit{``people and campfire"}, although the last prompt is most aligned with the \textit{``campfire"}. Our grounding module generates bounding boxes for the \textit{``campfire"}, resulting in a more aligned image with the specified context. \vspace{-0.0cm}}
\label{fig:ablation}
\end{figure*}

\noindent \textbf{Multimodal language model:}
Our work builds upon recent advancements in large-scale Transformer-based Language Models (LLMs). These models exhibit remarkable properties learning from few-shot in-context examples~\cite{brown2020language, chan2022transformers} and the ability to handle lengthy text inputs.  Some of the recent LLMs, like OpenAI's ChatGPT and GPT4~\cite{openai2023gpt}, have showcased impressive language comprehension and reasoning capabilities through techniques like instruction tuning~\cite{ouyang2022training, wu2023next, liu2023visual, zhang2023llavar} and reinforcement learning from human feedback (RLHF)~\cite{stiennon2020learning}. Moreover, a range of open-source LLMs, such as Flan-T5~\cite{chung2022scaling}, Vicuna~\cite{chiang2023vicuna}, LLaMA~\cite{touvron2023llama}, and Alpaca ~\cite{taori2023stanford}, have significantly accelerated progress and have made valuable contributions to the broader community.  Subsequently, there have been efforts to develop multimodal LLMs (MLLLMs) that can handle both multimodal inputs (image and text) and tasks. 

Most of the work in multimodal language models (MLLLMs)~\cite{huang2023language, zhu2023minigpt, su2022language, koh2023generating, wu2023next}, align pre-trained encoders of various modalities with the textual feature space of LLMs, allowing LLMs to effectively process other modal inputs, demonstrating compelling few-shot, captioning, and question-answering capabilities. Other approaches have built on this concept by introducing adapters ~\cite{eichenberg2021magma}, increasing model and data sizes ~\cite{alayrac2022flamingo}, improving visual encoders ~\cite{alayrac2022flamingo, li2023blip}, fine-tuning on instructions ~\cite{liu2023visual}, and training unified models with multi-task objectives ~\cite{lu2022unified, you2023cobit}. For example, Flamingo~\cite{alayrac2022flamingo} trained on 1535 TPUs for 15 days, while RA-CM3~\cite{yasunaga2023retrieval} utilized 256 GPUs for 5 days. Recent work, FROMAGe~\cite{koh2023grounding}, trained a multimodal LLM capable of processing arbitrarily interleaved image and text inputs to generate text interwoven with retrieved images. A closely related work to ours is GILL~\cite{koh2023generating}, which requires multimodal LLMs (MLLMs) to generate conditional embeddings, that are explicitly designed to align with a pre-trained CLIP encoder. These embeddings can subsequently be utilized with a pre-trained Stable Diffusion (SD) model ~\cite{rombach2022high}. 

\noindent\textbf{Text-to-Image Generation:}
The task of generating high-quality images \cite{Bhunia_2023_CVPR, kumar2021udbnet, kumar2023generative, kumar2023cross, NEURIPS2022_cd305fde} based on textual descriptions has gained popularity ~\cite{ramesh2022hierarchical, saharia2022photorealistic, rombach2022high}.  Latent Stable Diffusion ~\cite{rombach2022high} introduces denoising in the latent space and then decodes these denoised latents into high-resolution pixel space. However, These models fail to handle the complex and lengthy prompt. Recently, research studies like ~\cite{lian2023llm} and ~\cite{feng2023layoutgpt} have employed LLMs to tackle the challenges posed by lengthy text sequences. These methods use LLMs to generate layout using lengthy prompts.  %Different from these methods, along with the scene layout, we use semantic features learned by our alignment module to condition the diffusion model.

Given prior visual knowledge like layout, segmentation map, poses and stroke are used to condition to generate the novel image. ControlNet \cite{zhang2023adding}, GLIGEN \cite{li2023gligen}, and ReCo \cite{yang2023reco} have proposed training-based adaptations for spatially-conditioned image generation within the framework of diffusion models. However, these methods depend on external annotated datasets, like COCO \cite{lin2014microsoft}, which provide images with annotations such as bounding boxes and captions. Moreover, relying on training-based adaptation has dual implications. It not only results in the model's incompatibility because of add-ons such as pretrained LoRA \cite{hu2021lora} weights but also introduces complexities when attempting to train a new LoRA model as it requires additional training data to finetune. In contrast, \cite{lian2023llm} and \cite{feng2023layoutgpt} use training-free generation through existing text-to-image generation. Different from these methods which only condition the visual knowledge like layout to the existing text-to-image generator, we propose a cross-modal refinement module and layout generation using incontext examples to generate novel images.

\section{Method}
\label{sec:methodoogy}

\begin{figure*}[t!]
\centering
\includegraphics[width=1\textwidth]{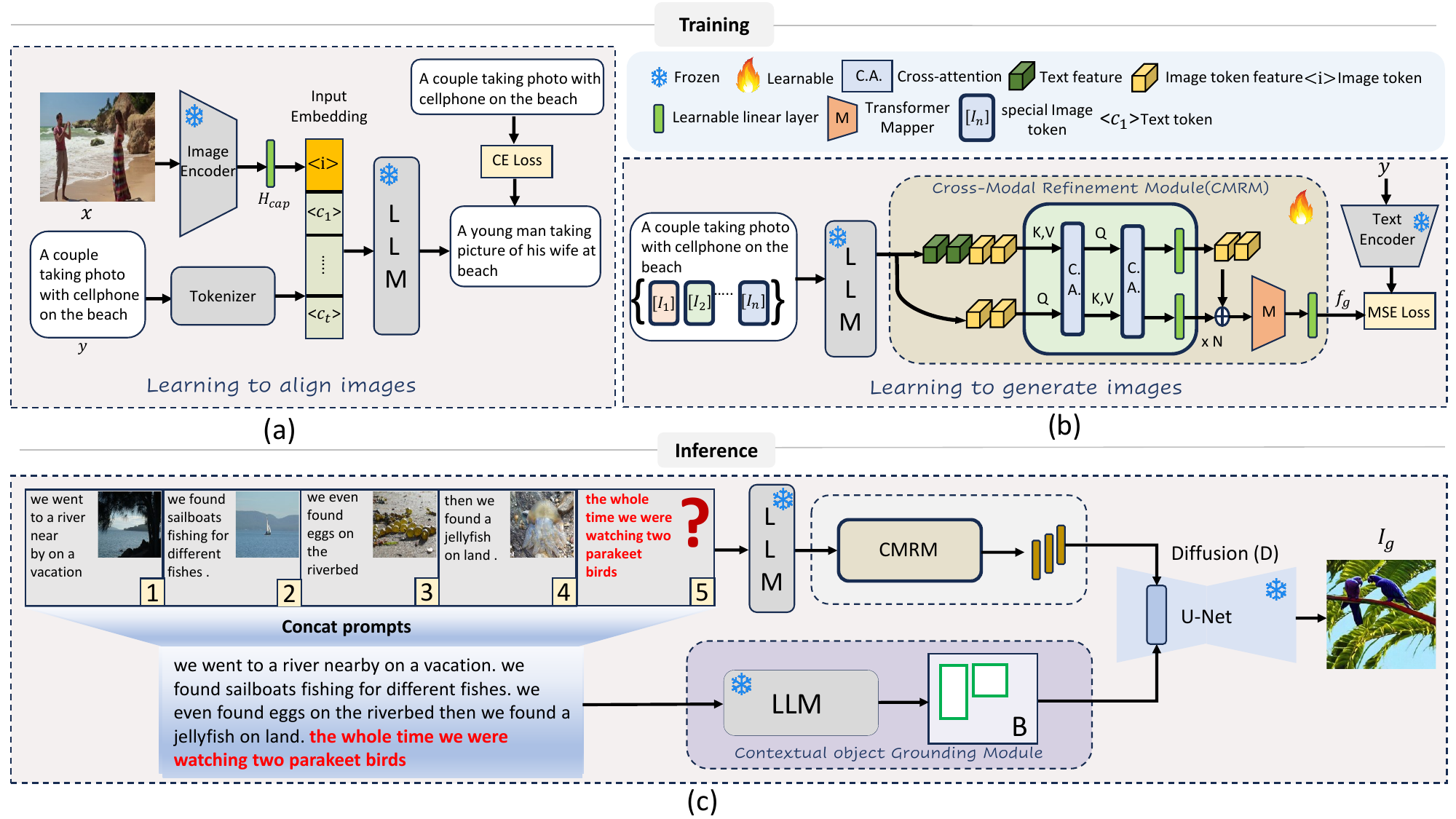}\vspace{0.0cm}
\caption{Overall framework of our model, MGCC to generate novel images using  multimodal prompts. During training,  \textbf{(a)} our model first align the image into the LLM token embedding space. \textbf{(b)}  To generate the novel images we introduce special image token $[I]$ to the LLM Vocabulary.  We refine these image token $[I]$ in the LLM feature space by introducing  a novel cross-modal refinement module (CMRM),  and then align these refined features in the clip text encoder space.   The refined image token $\hat{F}_{I}$ are then taken as input to the Transformer Mapper $S_w$ to map the tokens into the clip text embedding space as $f_g(y)$. \textbf{(c) }During inference, we use  Contextual Object Grounding, to generate the bounding boxes for the objects present in the scene $\{b_i\}_{i=1}^{p}$. We condition theses bounding boxes $\{b_i\}_{i=1}^{p}$ along with refined image tokens embedding $f_g$ on the diffusion $D$ to generate the final image $I_g$. }
\label{fig:Architecture}
% \vspace{-1em}
\end{figure*}

\textbf{Problem Statement:} 
%We aim to generate novel images using multi-modal language models (MLLM).  
Given a sequence of text prompts or interleaved image-text prompts (eg. story sequence) presented over multiple instances from $t_1$ to $t_{n-1}$ (\cref{fig:ablation}), our task is to generate the image at time $t_n$ while maintaining the context of the earlier text and image  prompt sequence. Here, $n$ represents the length of the sequence. Formally, our aim is to process interleaved sequences of text $y = \{y_a\}_{a=1}^n$ and images $x = \{x_{a}\}_{a=1}^{n-1}$   pairs, where $y$ and $x$ represent text and image respectively. Then, our objective is to generate a novel image at time $t_n$, retaining the context of earlier prompts. In this work, we leverage the capabilities of pre-trained and frozen large language models \cite{ouyang2022training, wu2023next, liu2023visual} and diffusion models \cite{ramesh2022hierarchical, rombach2022high} to generate these images with minimal training efforts. \\
\noindent\textbf{Baseline:} Our method builds upon the recent GILL approach \cite{koh2023generating}. In contrast to conventional diffusion models utilizing clip text encoders \cite{rombach2022high}, GILL adopts a pre-trained LLM. This fusion of LLM with the diffusion model enables image generation within extensive multimodal input. In processing text-image sequences, GILL \cite{koh2023generating} initially transforms the image into LLM embedding space. Additionally, it introduces specialized image tokens within the LLM's vocabulary to represent the final image to be generated by the model. These image tokens are aligned with the clip text encoder through a learnable transformer module named GILLMapper. Subsequently, GILLMapper's output serves as input to the diffusion model \cite{koh2023generating} during inference.

While our baseline GILL enables using lengthy multimodal story sequences, it faces several limitations:
\textbf{(a)} The output of GILLMapper, serving as the pre-trained diffusion model's input, tends to generate holistic images representing all prompts in a story sequence. This results in the \textit{loss of fine-grained details} specific to the current prompt at $t_n$. For instance, as shown in (\cref{intro-image} row 1), the baseline generates a holistic image combining information from various prompts, such as party foods and fruits, even when the final prompt corresponds solely to a pasta salad. \textbf{(b)} With increasing sequence length, GILL struggles to maintain \textit{coherent narrative and context}, evident in its inability to generate elephants in the final image (\cref{intro-image} row 2). 
\textbf{(c)} Performance deteriorates, particularly with lengthy descriptions and scenes featuring \textit{multiple objects}, impacting accurate image generation in such scenarios.

\noindent\textbf{Motivation:} As mentioned earlier, our baseline GILL introduces the image tokens within the vocabulary of a pretrained Language Model (LLM) to handle complex generation problems such as story sequence. While pretrained language models (LLM) are designed to capture dependencies within sequences of tokens, they are not explicitly optimized for capturing cross-modal relationships between text and special image tokens. In order to address this limitation and establish multi-model dependencies, we introduce a cross-modal refinement module (\cref{fig:Architecture}). This module enables the model to explicitly attend to relevant parts of the input when generating text and image tokens. 
Our proposed refinement is based on cross-attention and aims to refine the image token within LLM vocabulary such that the diffusion model does not generate a holistic image corresponding to all the prompts in the story sequence. It can produce images that contain both the semantics of the last prompt and the context of previous prompts of the sequence. 

Although the refinement of  image tokens improve the semantic understanding of the scene (\cref{fig:ablation}), the model  still lag behind the fine-grained understanding of the objects and their counts. %In Fig. \ref{fig:ablation}, row 1 the refined image tokens generate the image which has details of the objects  To further refine,
To address this, we introduce grounding of contextual objects with LLMs, to predict the layout of objects present in the last sequence of the story. By doing so, we are solving not only the problem of temporal consistency but also the problem of missing objects' in the scene and their inaccurate count, where both baseline GILL and the diffusion models are sub-optimal.   \cite{koh2023generating, li2023gligen, feng2023layoutgpt}.

\subsection{Overall Framework}

For a given training image $x$ and its caption $y$, we first perform an image  alignment as shown in \cref{fig:Architecture}  (a).  While, the input caption $y$ are tokenized as $(c_1 \cdots c_T)$,  the input image $x$ is passed through a pretrained clip visual encoder $g_{\phi}(x)$ to obtain the image embedding $g_{\phi}(x) \in \mathbb{R}^{d}$. Here $d$ is the dimension of the embeddings.  
The goal is to map these image embeddings $g_{\phi}(x)$ into a sequence of $k$ $e$-dimensional vectors, which serve as inputs to the pretrained LLMs. Here, $e$ represents the embedding dimension of the LLM. We  learn a linear mapping $H_{cap} \in \mathbb{R}^{d \times k_e}$ using the given image $x$ and captions $y$, to translate the $x$ into the token embedding space of the LLMs. This results in a mapping between the CLIP vision encoder and LLM (\cref{fig:Architecture} (a)).% is learned using image-caption pairs.   

To further enable the LLM to generate image outputs, a special set of tokens, named image tokens $ [I]=\text{\texttt{[I\{1\}]}}, \ldots, \text{\texttt{[I\{n\}]}}$ are introduced into the vocabulary of the pretrained LLMs as shown (\cref{fig:Architecture} (b)). Here, the  image tokens correspond to images that the model should generate as in  \cite{zhou2022learning, koh2023grounding, koh2023generating}.   The embedding matrix of LLMs, which maps words or tokens to continuous vector representations, is enhanced with an additional trainable matrix  $\mathbf{Emd} \in \mathbb{R}^{n \times e}$. This trainable matrix allows the model to better incorporate the specific characteristics to the final generated image $I_g$. % visa these image tokens $\text{\texttt{[I\{1\}]}}, \ldots, \text{\texttt{[I\{n\}]}}$.

We further introduce a Cross Modal Refinement Network that explicitly learn the cross-modal alignment to get the refined image token. Theses refined image token features obtained from the LLM  are further aligned to the clip text encoder, which then serves as an input to a diffusion model $D$ conditioned on bounding boxes \cite{li2023gligen}.  
We use a $4-$layer encoder-decoder transformer $S_w$ with the learnable weights \cite{koh2023generating} to learn the clip alignment. The transformer $S_w$ is conditioned on the refined image tokens processed by the LLM and a learnable query embedding $(q_1, \cdots, q_L) \in  \mathbb{T}^{L \times m}$ to extract  $L$ features from LLM hidden states.   Here, $m$ is embedding length of the transformer and $L$ is the maximum sequence length of the Diffusion model $D$ (similar to DETR \cite{carion2020end} and BLIP2 \cite{li2023blip}).  %This enables the transformer to extract  $L$ features from LLM hidden states.
 During Inference (see Fig.~\cref{fig:Architecture} (c)), we introduce  a contextual object grounding module (COGM) to predict the bounding boxes which are used along with the clip aligned features   to condition  the diffusion model. 
Next, we describe our cross model refinement contextual object grounding  modules.

\subsection{Cross Modal Refinement Module}
As discussed before,  the frozen and pretrained LLMs  are not explicitly designed to understand the cross-modal relationship between text and distinct image tokens representing an image within the LLM embedding space.  This results in the loss of fine-grained details specific to the final prompt within the generated images.  For instance, in \cref{fig:ablation}, our baseline GILL model \cite{koh2023generating} generates \textit{``coffee and snack"} on the table which doesn't align with prompt \textit{``the boss is teaching the new employee to prepare the coffee and snack"} and also lose the context of the earlier prompts \textit{``like shop and customers"}.

To solve this problem, we introduce a refinement network that explicitly learns the cross-modal dependencies using the cross-attention between the special image token and the text in the LLM embedding space. To learn the refinement module, we pass text $y$ and the image tokens $[I]$ to the LLM and obtain a multimodal feature embedding $f_{mm}$ in the LLM space. This multimodal feature contains the embedding representation of both text and image tokens. We first separate the embedding of image tokens  $f_{I} \subset f_{mm}$ before applying cross-attention between $f_I$ and $f_{mm}$.

\begin{equation}
    \mathrm{Attn^{joint}} = \left( \frac{proj_{q,I}(f_{I})proj_{q,{mm}}(f_{mm})^T}{\sqrt{d^k}} \right),
    \label{eq:meth:encoder:2}
\end{equation}
 
where $proj_{q,I_n}$ and $proj_{q,y}$ are the query projections for the image token and text features.
\begin{equation}
\begin{aligned}
    &F_{I} = \mathrm{FFN}^{m}_{I}(\mathrm{softmax}(\mathrm{Attn^{joint}})proj_{t}(f_{mm}))), \\
    &F_{mm} = \mathrm{FFN}^{m}_y(\mathrm{softmax}(\mathrm{Attn^{joint}}^T)proj_{I}(f_{I}))),
    \label{eq:meth:encoder:3}
\end{aligned}
\end{equation}
\noindent where $\mathrm{FFN}$ denotes learnable linear layer, $F_{I}$ and $F_{mm}$ are refined features by our refinement module. Then, we apply the following operation to obtain the final image tokens. 

\begin{equation}
\begin{aligned}
    & \hat{F_{I}} = (F_{mm} \odot m_I)  + F_{I},
    \label{eq:meth:encoder:4}
\end{aligned}
\end{equation}
\noindent where $m_I$ is a mask with $0$s for the text tokens and $1$s for the image tokens. 
Finally, we pass the refined image tokens $\hat{F_{I}}$ to transformer $S_w$ to align them to the clip text encoder space, 
% 
%further we align these refine image token $\hat{F_{I_n}}$ using the similar transformer architecture $S_w$
% \cite{koh2023generating}.  
\begin{equation}
\begin{aligned}
  & f_g(y) = s_w(\hat{F_{I_1}}, \hat{F_{I_2}}, \cdots, \hat{F_{I_n}}, (q_1, \cdots, q_L)).
\end{aligned}
\end{equation}
Here $f_g \in \mathbb{R}^{1 \times L}$ is the output of the transformer. 
%We train this network with the same losses as our baseline \cite{koh2023generating}.

\subsection{ Contextual Object Grounding Module}
% \noindent \textbf{ Contextual Object Grounding Module:}
 We perform in-context learning during inference to generate images that captures fine-grained details within all prompt  sequences. Although the cross-modal refinement module improves the semantics, the generated images  still lack behinds several aspects. For example, in \cref{fig:ablation},  the model fails  to generate relevant objects present in the previous prompts of this sequence.   During the in-context learning, our  contextual object grounding module detects the \textit{``campfire"} in the whole image and generates it accordingly as shown in our final results.

Specifically,  during inference, given a story sequence $(y_1, y_2 \cdots y_n)$, we generate the bounding boxes $b_1 \cdots b_N$ of relevant objects along with their class labels. 
Here, $b_i = [x, y, w, h]$ where $w, h$ are the height and width of the bounding boxes.   These bounding boxes are obtained from the LLMs thorough specific prompting explained below.

\noindent\textbf{Prompting}.
We designed a prompt for LLMs as follows:

\begin{enumerate}[leftmargin=*]
\item {Description of the task:}
\begin{tcolorbox}[colback=gray!5!white,colframe=gray!75!black,boxsep=2pt,left=2pt,right=2pt,top=2pt,bottom=2pt]
\textit{You are an intelligent bounding box generator. I will provide you with a entire story sequence for a occasion. Your task is to generate the bounding boxes for the last sequence remaining the context of the earlier sequence}
\end{tcolorbox}
\item {Detail of the image:}
\begin{tcolorbox}[colback=gray!5!white,colframe=gray!75!black,boxsep=2pt,left=2pt,right=2pt,top=2pt,bottom=2pt]
\textit{The images are of size $512\times 512$ ... Format of the bounding boxes should be fixed ... If needed, take the context of the previous sequence and have the guess}.
\end{tcolorbox}
\end{enumerate}

Similar to the \cite{lian2023llm, brooks2023instructpix2pix}, we prompt the Large Language Model (LLM) with manually curated context examples subsequent to predict the bounding boxes. These examples serve to elucidate the layout representation and help eliminate potential ambiguities. An example is provided below:

\begin{tcolorbox}[colback=gray!5!white,colframe=gray!75!black,boxsep=2pt,left=2pt,right=2pt,top=2pt,bottom=2pt]
\textit{Story sequence: We took my son on a roadtrip. We stopped to look at the golden gate bridge. He had a lot of fun in the go carts. We stopped in the desert and took a picture. He was excited to get home. \\
Objects: [(`a car', [482, 100, 27, 18]), (`a child', [102, 107, 201, 402])] 
}
\end{tcolorbox}

The LLM generates the bounding boxes for the last prompt in the sequence  as we can see in the  above example where both \textit{``car"} and \textit{``child"} are  not mentioned in the last prompt but LLMs need to have the understanding of the previous sequence to predict the final content in the scene.
The bounding boxes $B$ and features learned by our alignment network $f_g$ are then passed through the Diffusion model $D$ to generate the final image $I_g$ which contains the semantics as well as the fine-grained details about the sequence.% in context of multimodal input. 

\section{Experiment}
\label{sec:experiment}

\textbf{Datatset:} The proposed MGCC method is evaluated on two datasets: 
Visual Story Generation (VIST)  and Visual Dialogue Context (VisDial). \\
\noindent\textbf{VIST \cite{huang2016visual}}: The VIST dataset contains a collection of sequences for vision-and-language tasks, featuring 5 text-image pairs that form a cohesive story. Similar to \cite{koh2023grounding, koh2023generating}, our evaluation is performed by generating the final image in the sequence of texts under three different input conditions: (a) \emph{Single Caption}: The input comprises of only the last text description. This scenario mirrors standard text-to-image generation, where the model is conditioned on a single caption to produce an image. (b) \emph{Multiple Captions (5 captions)}: The input encompasses text descriptions from the entire story sequence. This assessment evaluates the models' capability to handle longer and temporally dependent text descriptions. (c) \emph{Multimodal Context (5 Captions, 4 Images)}: The input encompasses all the image-text pairs preceding the final image, and additionally the last text description. This evaluation assesses the models' proficiency in processing \textit{multimodal context} during image generation.  \textbf{VisDial ~\cite{das2017visual}} contains sequences of question-answer (Q$\&$A) pairs related to a specific image that simulate a dialogue between two individuals discussing the image. Each example incorporates up to 10 rounds of Q$\&$A pairs. This evaluates the model's generalizability to dialogue-like text and its ability to process long-text sequences.
% \end{enumerate}

\noindent\textbf{Evaluation Metrics:}
Our evaluation focuses on assessing the capability of the model to handle complex prompt descriptions. To measure the relevance of the generated image content, we employ two standard evaluation metrics, CLIP Similarity and LPIPS. \textbf{CLIP Similarity} utilizes the CLIP~\cite{radford2021learning} ViT-L image encoder, to extract feature representations for both the real and generated images, and calculates the cosine similarity between them. A higher score indicates greater similarity.
\textbf{Learned Perceptual Image Patch Similarity (LPIPS)} ~\cite{zhang2018unreasonable} measures the distance between image patches, assessing the dissimilarity between real and generated images. A lower LPIPS value signifies a closer perceptual resemblance, while a higher value indicates greater dissimilarity.

\begin{figure}
    \centering
    % Row 1: Figure 1

    %\vspace{-3pt}
    \begin{minipage}[b]{0.4\columnwidth}
        \centering
         \includegraphics[width=0.83\columnwidth]{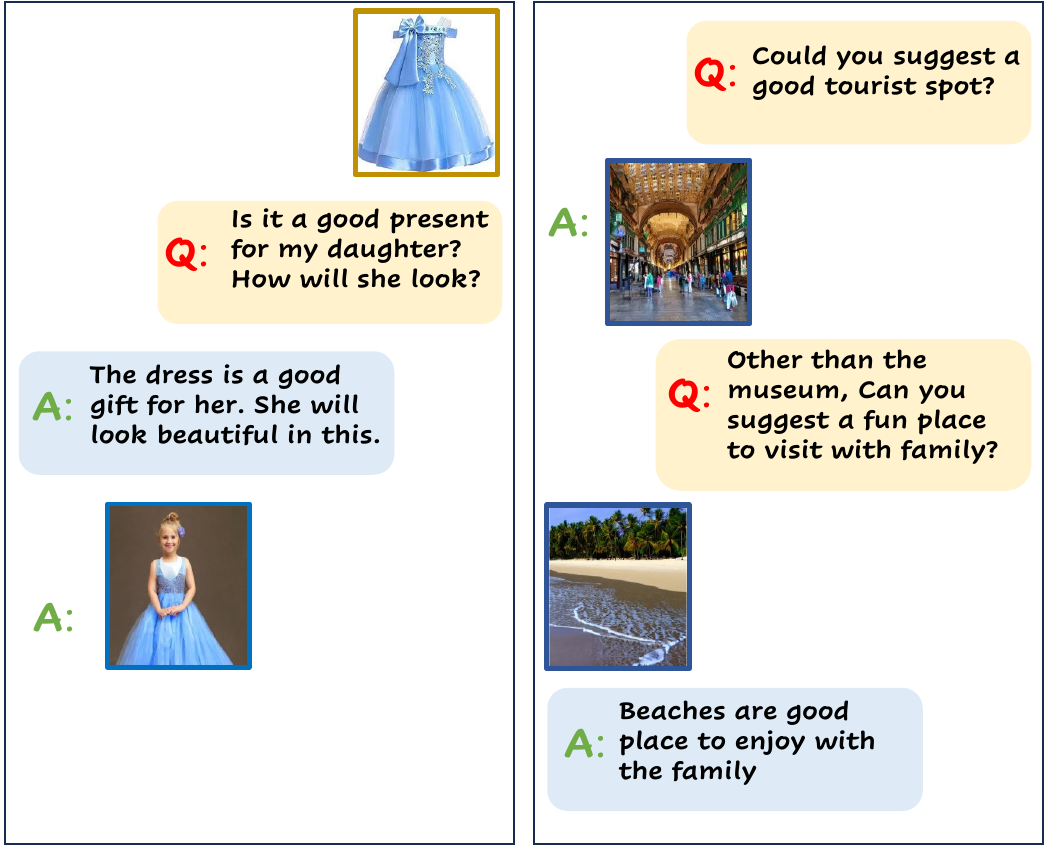}
          % \vspace{-25pt}
        % \subcaption{}
        % \vspace{-15pt}
        \label{fig:plug}
    \end{minipage}%
    % Row 1: Table 1
    \begin{minipage}[b]{0.6\columnwidth}
        \centering
         \includegraphics[width=1.0\columnwidth]{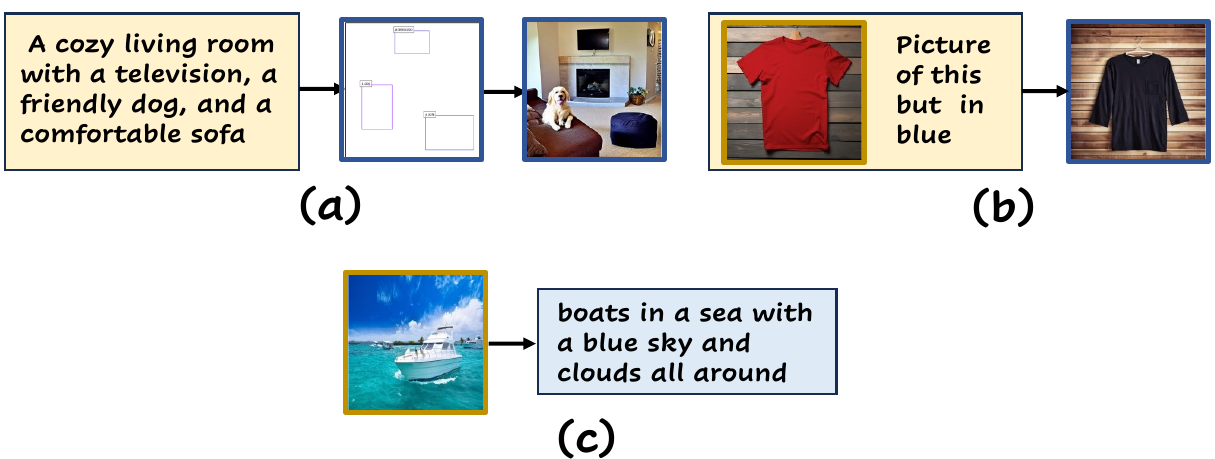}
          % \vspace{-25pt}
        % \subcaption{}
        % \vspace{-15pt}
        \label{fig:multimodal}
    \end{minipage}%
    \caption{The images on the left showcase examples illustrating the multimodal generation capabilities of our MGCC, which operates on sequential multimodal input dialogues arranged from top to bottom.  On the right-hand side, the images demonstrate: (a) the model's ability to perform grounded generation, (b) its proficiency in following instructions, and (c) its capability in generating descriptive captions for images.}
    \label{fig:multimodal}
    \vspace{-10pt}
\end{figure}

\subsection{Implementation Details}
% \textbf{Implementation Details:} 
Following \cite{koh2023generating, koh2023grounding}, we train on the Conceptual Captions (CC3M) dataset~\cite{sharma2018conceptual} comprising of 3.3 million image-text pairs. The OPT-6.7B model~\cite{zhang2022opt} serves as the language model with hidden state embedding dimension $e=4096$. To align the input image in the LLMs token embedding space we employ CLIP~\cite{radford2021learning} ViT-L image encoder. For the text-to-image generation module ($D$) we employ Gligen~\cite{li2023gligen} backbone network. All the weights from the pre-trained models are kept frozen, updating only the linear layer $\mathbf{H}_{\text{cap}}$, embedding matrix $\mathbf{Emb}_{\text{img}}$, cross-modal refinement module and the transformer mapper $\mathbf{S_w}$. Similar to \cite{koh2023generating}, we use $k = 4$ visual tokens and $n = 8$ learned $[I]$ tokens. The embedding dimension of the learnable query $q$ is set to $m=512$. The total number of refinement layers is set to $4$. We optimize using Adam~\cite{kingma2014adam} with a learning rate of 0.001 and parameters $\beta_1 = 0.9$ and $\beta_2 = 0.95$. The total number of in-context examples to generate the image bounding boxes used during inference is set to $5$. We train this network with the same losses as our baseline \cite{koh2023generating} including the cross entropy (CE)  and the mean squared  (MSE) losses.

\begin{table*}[t]
  \caption{Performance comparison with existing approaches on VIST \cite{huang2016visual}. For single caption inputs, compared to Stable Diffusion, our approach performs on par in terms of CLIP similarity, while performing favorably in terms of LPIPS. Furthermore, our MGCC outperforms both Stable Diffusion and GILL, in terms of both metrics for long sequence of captions (5 captions) and multimodal inputs (5 caps, 4 images), highlighting our approach's improved alignment in generating context-aware images while maintaining temporal consistency.\vspace{-0.0cm}}
  \label{tab:vist_generation}
  \centering
  \resizebox{1.0\linewidth}{!}{%
  \begin{tabular}{lcccccccc}
    \toprule
    &&   \multicolumn{3}{c}{\textbf{CLIP Similarity} ($\uparrow$)} &&  \multicolumn{3}{c}{\textbf{LPIPS} ($\downarrow$)}      \\
    \cmidrule(r){3-5} \cmidrule(r){7-9}
    \textbf{Model}    && 1 caption     & 5 captions & 5 caps, 4 images  &&  1 caption     & 5 captions & 5 caps, 4 images \\
    \midrule
    GLIDE~\cite{nichol2021glide} &&  0.582  &  0.591 & -  &&  0.753  &  0.745 & -   \\
    % {\color{red}{DeepFloyd}~\cite{}} &&  ?  &  ? &  -  &&  ?  &  ? &  -   \\
    Stable Diffusion~\cite{rombach2022high} &&  \textbf{0.592} $\pm 0.0007$  &  $0.598 \pm 0.0006$ & -  &&  0.703 $\pm 0.0003$  &  $0.704 \pm 0.0004$ & -   \\ 
    GILL \cite{koh2023generating}   &&  0.581 $\pm 0.0005$  &  0.612  $\pm 0.0011$ & 0.641 $\pm 0.0011$  &&  0.702 $\pm 0.0004$  &  0.696 $\pm 0.0008$  &   0.693 $\pm 0.0008$  \\
    \midrule
    \cellcolor{cyan!25}\textbf{Ours: MGCC}   &\cellcolor{cyan!25}\cellcolor{cyan!25}&  \cellcolor{cyan!25}0.591 $\pm 0.0002$ & \cellcolor{cyan!25} \textbf{0.637} $\pm 0.0007$ & \cellcolor{cyan!25} \textbf{0.652} $\pm 0.0009$  &\cellcolor{cyan!25}& \cellcolor{cyan!25} \textbf{0.699} $\pm 0.0015$  &  \cellcolor{cyan!25} \textbf{0.682} $\pm 0.0018$  & \cellcolor{cyan!25}  \textbf{0.679} $\pm 0.0012$ \\
    \bottomrule
  \end{tabular}
  }
  % \vspace{-0.2in}
\end{table*}

\begin{figure*}[t!]
\centering
\includegraphics[width=1\textwidth]{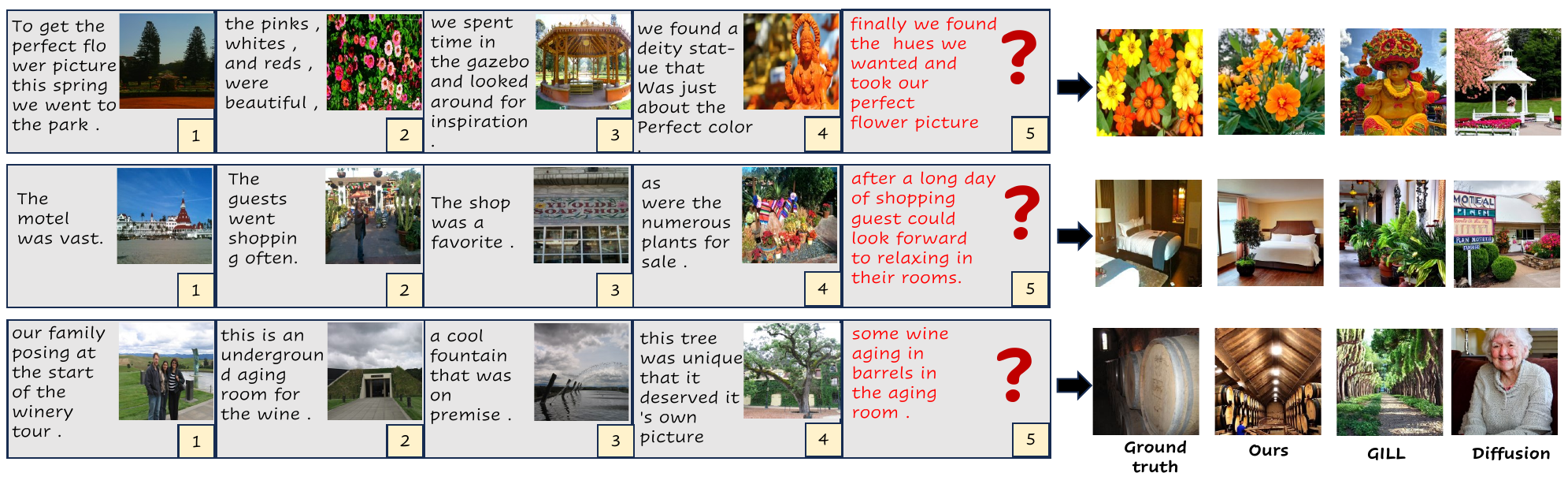}\vspace{-0.0cm}
\caption{In the  first row, the baseline model produces a holistic representation of the scene, including the \textit{``statue"} and the \textit{``flowers"}. However, our model excels in generating \textit{``hue"} flower picture. In the second row, the baseline model fails to comprehend the context sequence about the \textit{``room"} and the \textit{``guest"}, whereas our model successfully captures this context, resulting in generating the \textit{``room having the bed"} with the help of the context of \textit{``relaxing"}. Moving to the third row, the baseline and diffusion loses the context as the prompt sequence increases and generates \textit{``trees"} and \textit{``old lady"} whereas our model can generate the images very much aligned with the text \textit{``barrels in the aging room"} groundtruth.   \vspace{-0.0cm}
}
\label{fig:Qual_VIST}
\end{figure*}

\subsection{Experimental Results}
\begin{figure*}[t!]
\centering
\includegraphics[width=1\textwidth]{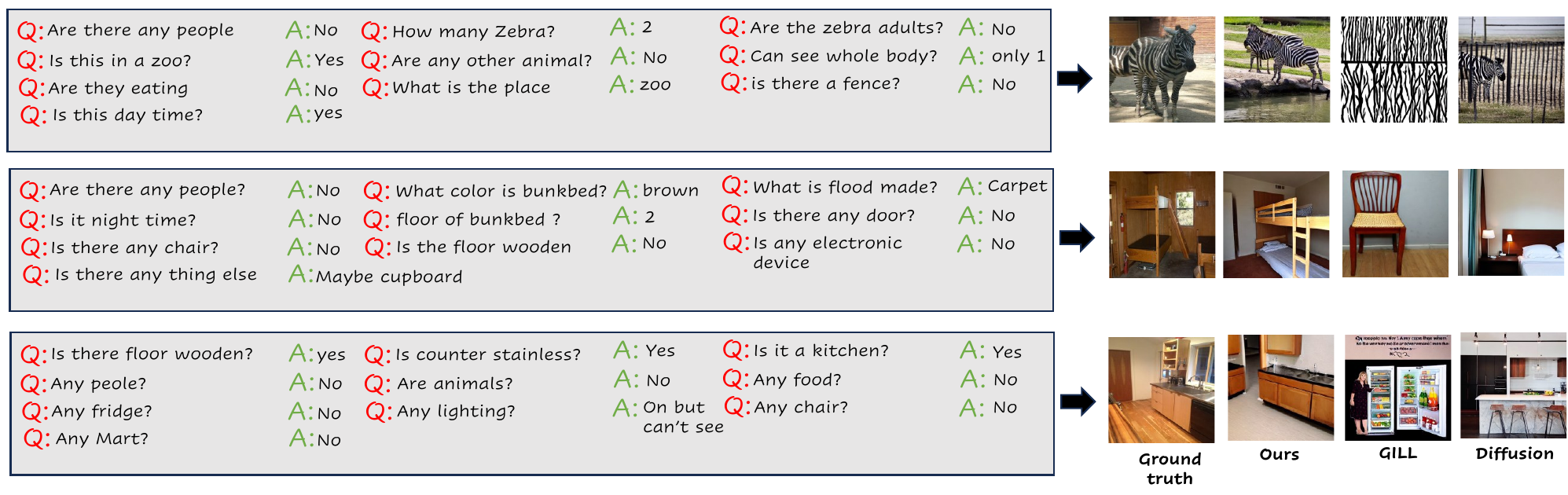}\vspace{-0.0cm}
\caption{In the first example, baseline and diffusion models gets confused between the \textit{``fence"} and \textit{``zebra"} whereas our model is able to get the \textit{``two zebra"} with fine grained details like \textit{``can see whole body of only one"}. In the second case, the baseline and diffusion model failed to comprehend the scene and only generated the \textit{``chair"} and a \textit{``bed"} as its output. In contrast, our model demonstrated its capability by generating the image of \textit{``bunky beds"} with fine grained details like \textit{floor of bunkbed}. In the third row, the baseline and diffusion fails to generate the \textit{wooden floor kitchen} whereas our model is able to generate image aligned with the ground truth.\vspace{-0.0cm}
}
\label{fig:Qual_Visdial}
\end{figure*}

% \noindent \textbf{Experimental Results} \\
\textbf{Quantitative and Qualitative Results:}
We present the quantitative comparison of our proposed approach MGCC on datasets VIST \cite{huang2016visual} and VisDial \cite{das2017visual} in \cref{tab:vist_generation} and  \cref{tab:visdial_generation} respectively. In  \cref{tab:vist_generation} we observe that when a single caption is provided, our model's performance closely aligns with stable diffusion \cite{rombach2022high}, while marginally outperforming the baseline GILL. However, when a sequence of 5 captions from the story is given as input, our model surpasses both stable diffusion and GILL, improving CLIP Similarity from $0.612$ to $0.637$ and LPIPS from $0.696$ to $0.682$. Further investigation with multimodal story sequences (5 captions and 4 images) improves the CLIP Similarity score from $0.641$ to $0.652$ and LPIPS from $0.693$ to $0.679$. This demonstrates that our model is capable of capturing multimodal inputs and a sequence of lengthy prompts to generate images that are contextually aligned, and the qualitative results for the same can be seen in \cref{fig:Qual_VIST}.
In \cref{tab:visdial_generation}, we observe that for short rounds of dialogue, stable diffusion outperforms both GILL and MGCC. However, for long dialogues sequences, MGCC outperforms both GILL and stable diffusion, improving CLIP Similarity from $0.645$ to $0.660$ and LPIPS from $0.714$ to $0.699$.  These results indicate that our model is able to handle the lengthy prompts sequence and dialogue-like inputs better. As shown in \cref{fig:Qual_Visdial} MGCC demonstrates a keen understanding of objects and their count within the dialogue. This could be attributed to the cross-modal refinement module which enhances the image tokens for better semantics, and our contextual object grounding module contributes to generating fine-grained details in the images. Our model MGCC, can process multimodal dialogue to generate multimodal (images and text) outputs as shown in Fig.~\ref{fig:multimodal}.

\begin{table*}[t]
  \caption{Performance comparison with existing approaches on the VisDial dataset \cite{das2017visual}, in terms of CLIP similarity and LPIPS. While Stable Diffusion performs favorably for short rounds of dialogue, our MGCC approach outperforms both Stable Diffusion and GILL for long dialogue sequences (5 rounds, 10 rounds), indicating that our approach handles lengthy prompts and dialogue-like inputs better.\vspace{-0.0cm}}
  \label{tab:visdial_generation}
  \centering
  \resizebox{1.0\linewidth}{!}{ %
  \begin{tabular}{lcccccccc}
    \toprule
    &&   \multicolumn{3}{c}{\textbf{CLIP Similarity} ($\uparrow$)} &&  \multicolumn{3}{c}{\textbf{LPIPS} ($\downarrow$)}      \\
    \cmidrule(r){3-5} \cmidrule(r){7-9}
    \textbf{Model}    && 1 round     & 5 rounds & 10 rounds  && 1 round     & 5 rounds & 10 rounds \\
    \midrule
    GLIDE~\cite{nichol2021glide} &&  \textbf{0.562}  &  0.595 & 0.587  &&  0.800  &  0.794 & 0.799   \\
    % {\color{red}{DeepFloyd}~\cite{}} &&  ?  &  ? &  -  &&  ?  &  ? &  -   \\
    Stable Diffusion~\cite{rombach2022high} &&  0.552 $\pm 0.0015$  &  $0.629 \pm 0.0015$ & 0.622 $\pm 0.0012$  &&  \textbf{0.642} $\pm 0.0010$  &  $0.722 \pm 0.0012$ & $0.723 \pm 0.0008$   \\ 
    GILL \cite{koh2023generating}   &&  0.528 $\pm 0.0014$  &  0.621  $\pm 0.0009$ & 0.645 $\pm 0.0010$  &&  0.742 $\pm 0.0004$  &  0.718 $\pm 0.0028$  &   0.714 $\pm 0.0006$  \\
    \midrule
    \cellcolor{cyan!25} \textbf{Ours: MGCC}   &\cellcolor{cyan!25}& \cellcolor{cyan!25} 0.539 $\pm 0.0009$ & \cellcolor{cyan!25} \textbf{0.639} $\pm 0.0010$ & \cellcolor{cyan!25}\textbf{0.660} $\pm 0.0003$ &\cellcolor{cyan!25}& \cellcolor{cyan!25} 0.712  $\pm 0.0019$   &  \cellcolor{cyan!25} \textbf{0.704}  $\pm 0.0015$ & \cellcolor{cyan!25}  \textbf{0.699} $\pm 0.0012$ \\
    \bottomrule
  \end{tabular}
  }
  % \vspace{-0.2in}
\end{table*}

\begin{figure*}[h!]
\centering
\begin{minipage}{0.5\textwidth}
\centering
\captionof{table}{Image generation performance on CC3M~\cite{sharma2018conceptual} and VIST~\cite{huang2016visual} with our proposed contribution onto the baseline.}% \TextAdapter{} substantially outperforms other baseline mapping networks.}
\vspace{-0.00in}
\label{tab:ablation_1}
\setlength{\tabcolsep}{12pt}
\resizebox{\textwidth}{!}{%

\begin{tabular}{lcccc}
\toprule
    &&       \textbf{CC3M}   &&  \textbf{VIST}  \\
\cmidrule{3-3}  \cmidrule{5-5}
  \textbf{Model}  &&   \textbf{FID} ($\downarrow$)  &&  \textbf{CLIP Sim} ($\uparrow$)  \\
\midrule
Stable Diffusion~\cite{rombach2022high}  && \textbf{13.94} && 0.598 \\

Baseline    &&   15.31   &&  0.641  \\
\midrule
Baseline + COGM &&     14.98    && 0.644  \\
Baseline + CMRM &&     14.67    &&  0.646 \\
\cellcolor{cyan!25}Ours(Baseline + COGM + CMRM) &\cellcolor{cyan!25}&  \cellcolor{cyan!25}   14.23   &\cellcolor{cyan!25}& \cellcolor{cyan!25} \textbf{0.652}\\
\bottomrule
\end{tabular}
}
\end{minipage}
\hfill% Add a space between the two minipages
\begin{minipage}{0.37\textwidth}
\centering
\captionof{table}{Image generation result on different number of layer of cross modal refinement module (CMRM) .}
\vspace{-0.00in}
\label{tab:image_token_ablation}
\setlength{\tabcolsep}{10pt}
\resizebox{\textwidth}{!}{%

\begin{tabular}{lcccc}
\toprule
&& \textbf{CC3M} && \textbf{VIST} \\
\cmidrule{3-3}  \cmidrule{5-5}
\textbf{$N$} \hspace{3mm}    &&    \textbf{FID} ($\downarrow$)    && \textbf{CLIP Sim} ($\uparrow$) \\
\midrule
1     &&    15.11  &&  0.643 \\
2     &&    14.83    &&   0.6470\\
\cellcolor{cyan!25}4     &\cellcolor{cyan!25}&  \cellcolor{cyan!25}  \textbf{14.23}    &\cellcolor{cyan!25}& \cellcolor{cyan!25} \textbf{0.652} \\
\bottomrule
\end{tabular}
}
\end{minipage}%
% \vspace{-0.1in}
\end{figure*}

% \vspace{-1.5em}
\noindent\textbf{Ablation Study:}

Here, we present the impact of the two proposed modules: cross-modal refinement (CMRM) and contextual object grounding (COGM) on the CC3M~\cite{sharma2018conceptual} and VIST~\cite{huang2016visual} datasets in \cref{tab:ablation_1}.
% We can observe progressive improvements on the baseline with the addition of each module. 
When integrating the CMRM and COGM into the baseline, we observe progressive improvement in FID and CLIP Similarity scores. This highlights the importance of learning the cross-modal dependencies across two different modalities (image and text) and learning fine-grained details of the objects. Finally, our proposed approach, which simultaneously integrates the two modules takes advantage of both these capabilities to generate fine-grained semantically aligned images. This is reflected in the improvements in the FID from $15.31$ to $14.23$ and CLIP Similarity from $0.641$ to $0.652$. In \cref{tab:image_token_ablation}, we ablate the impact of the number of cross-modal refinement module in MGCC. We observe that with the increase in the number of modules, the FID improves from $15.11$ to $14.23$ and CLIP Similarity from $0.643$ to $0.652$ for module $N=1$ to $N=4$ respectively. This indicates the models' ability to capture improved cross-modal dependencies. % We observe that with the increase in the number of modules, the FID improves from $15.11$ 

\subsection{Conclusion}

We present MGCC, a method designed to generate images from lengthy and complex multimodal prompt sequences while maintaining temporal consistency. Our approach involves a cross-modal refinement module explicitly learning correspondence between multimodal inputs (image and text) and integrates contextual object grounding for precise control of object layout and count. Quantitative and qualitative results on two benchmark datasets demonstrate the merits of our contributions. On both datasets, our method demonstrates superior image generation quality and alignment with ground truth compared to existing approaches. 
\clearpage  
\bibliographystyle{splncs04}
\bibliography{main}

\begin{thebibliography}{10}
\providecommand{\url}[1]{\texttt{#1}}
\providecommand{\urlprefix}{URL }
\providecommand{\doi}[1]{https://doi.org/#1}

\bibitem{aghajanyan2022cm3}
Aghajanyan, A., Huang, B., Ross, C., Karpukhin, V., Xu, H., Goyal, N., Okhonko, D., Joshi, M., Ghosh, G., Lewis, M., et~al.: Cm3: A causal masked multimodal model of the internet. arXiv preprint arXiv:2201.07520  (2022)

\bibitem{ahn2022can}
Ahn, M., Brohan, A., Brown, N., Chebotar, Y., Cortes, O., David, B., Finn, C., Fu, C., Gopalakrishnan, K., Hausman, K., et~al.: Do as i can, not as i say: Grounding language in robotic affordances. arXiv preprint arXiv:2204.01691  (2022)

\bibitem{alayrac2022flamingo}
Alayrac, J.B., Donahue, J., Luc, P., Miech, A., Barr, I., Hasson, Y., Lenc, K., Mensch, A., Millican, K., Reynolds, M., et~al.: Flamingo: a visual language model for few-shot learning. Advances in Neural Information Processing Systems  \textbf{35},  23716--23736 (2022)

\bibitem{Bhunia_2023_CVPR}
Bhunia, A.K., Koley, S., Kumar, A., Sain, A., Chowdhury, P.N., Xiang, T., Song, Y.Z.: Sketch2saliency: Learning to detect salient objects from human drawings. In: Proceedings of the IEEE/CVF Conference on Computer Vision and Pattern Recognition (CVPR). pp. 2733--2743 (June 2023)

\bibitem{brooks2023instructpix2pix}
Brooks, T., Holynski, A., Efros, A.A.: Instructpix2pix: Learning to follow image editing instructions. In: Proceedings of the IEEE/CVF Conference on Computer Vision and Pattern Recognition. pp. 18392--18402 (2023)

\bibitem{brown2020language}
Brown, T., Mann, B., Ryder, N., Subbiah, M., Kaplan, J.D., Dhariwal, P., Neelakantan, A., Shyam, P., Sastry, G., Askell, A., et~al.: Language models are few-shot learners. Advances in neural information processing systems  \textbf{33},  1877--1901 (2020)

\bibitem{carion2020end}
Carion, N., Massa, F., Synnaeve, G., Usunier, N., Kirillov, A., Zagoruyko, S.: End-to-end object detection with transformers. In: European conference on computer vision. pp. 213--229. Springer (2020)

\bibitem{chan2022transformers}
Chan, S.C., Dasgupta, I., Kim, J., Kumaran, D., Lampinen, A.K., Hill, F.: Transformers generalize differently from information stored in context vs in weights. arXiv preprint arXiv:2210.05675  (2022)

\bibitem{chen2023minigptv2}
Chen, J., Zhu, D., Shen, X., Li, X., Liu, Z., Zhang, P., Krishnamoorthi, R., Chandra, V., Xiong, Y., Elhoseiny, M.: Minigpt-v2: large language model as a unified interface for vision-language multi-task learning. arXiv preprint arXiv:2310.09478  (2023)

\bibitem{chiang2023vicuna}
Chiang, W.L., Li, Z., Lin, Z., Sheng, Y., Wu, Z., Zhang, H., Zheng, L., Zhuang, S., Zhuang, Y., Gonzalez, J.E., et~al.: Vicuna: An open-source chatbot impressing gpt-4 with 90\%* chatgpt quality. See https://vicuna. lmsys. org (accessed 14 April 2023)  (2023)

\bibitem{chung2022scaling}
Chung, H.W., Hou, L., Longpre, S., Zoph, B., Tay, Y., Fedus, W., Li, Y., Wang, X., Dehghani, M., Brahma, S., et~al.: Scaling instruction-finetuned language models. arXiv preprint arXiv:2210.11416  (2022)

\bibitem{das2017visual}
Das, A., Kottur, S., Gupta, K., Singh, A., Yadav, D., Moura, J.M., Parikh, D., Batra, D.: Visual dialog. In: Proceedings of the IEEE conference on computer vision and pattern recognition. pp. 326--335 (2017)

\bibitem{NEURIPS2022_cd305fde}
Dong, Q., Muhammad, A., Zhou, F., Xie, C., Hu, T., Yang, Y., Bae, S.H., Li, Z.: Zood: Exploiting model zoo for out-of-distribution generalization. In: Koyejo, S., Mohamed, S., Agarwal, A., Belgrave, D., Cho, K., Oh, A. (eds.) Advances in Neural Information Processing Systems. vol.~35, pp. 31583--31598. Curran Associates, Inc. (2022), \url{https://proceedings.neurips.cc/paper_files/paper/2022/file/cd305fdee96836d5cc1de94577d71b61-Paper-Conference.pdf}

\bibitem{driess2023palm}
Driess, D., Xia, F., Sajjadi, M.S., Lynch, C., Chowdhery, A., Ichter, B., Wahid, A., Tompson, J., Vuong, Q., Yu, T., et~al.: Palm-e: An embodied multimodal language model. arXiv preprint arXiv:2303.03378  (2023)

\bibitem{eichenberg2021magma}
Eichenberg, C., Black, S., Weinbach, S., Parcalabescu, L., Frank, A.: Magma--multimodal augmentation of generative models through adapter-based finetuning. arXiv preprint arXiv:2112.05253  (2021)

\bibitem{feng2023layoutgpt}
Feng, W., Zhu, W., Fu, T.j., Jampani, V., Akula, A., He, X., Basu, S., Wang, X.E., Wang, W.Y.: Layoutgpt: Compositional visual planning and generation with large language models. arXiv preprint arXiv:2305.15393  (2023)

\bibitem{hu2021lora}
Hu, E.J., Shen, Y., Wallis, P., Allen-Zhu, Z., Li, Y., Wang, S., Wang, L., Chen, W.: Lora: Low-rank adaptation of large language models. arXiv preprint arXiv:2106.09685  (2021)

\bibitem{huang2023language}
Huang, S., Dong, L., Wang, W., Hao, Y., Singhal, S., Ma, S., Lv, T., Cui, L., Mohammed, O.K., Liu, Q., et~al.: Language is not all you need: Aligning perception with language models. arXiv preprint arXiv:2302.14045  (2023)

\bibitem{huang2016visual}
Huang, T.H., Ferraro, F., Mostafazadeh, N., Misra, I., Agrawal, A., Devlin, J., Girshick, R., He, X., Kohli, P., Batra, D., et~al.: Visual storytelling. In: Proceedings of the 2016 conference of the North American chapter of the association for computational linguistics: Human language technologies. pp. 1233--1239 (2016)

\bibitem{ilharco2020probing}
Ilharco, G., Zellers, R., Farhadi, A., Hajishirzi, H.: Probing contextual language models for common ground with visual representations. arXiv preprint arXiv:2005.00619  (2020)

\bibitem{kingma2014adam}
Kingma, D.P., Ba, J.: Adam: A method for stochastic optimization. arXiv preprint arXiv:1412.6980  (2014)

\bibitem{koh2023generating}
Koh, J.Y., Fried, D., Salakhutdinov, R.: Generating images with multimodal language models. NeurIPS  (2023)

\bibitem{koh2023grounding}
Koh, J.Y., Salakhutdinov, R., Fried, D.: Grounding language models to images for multimodal inputs and outputs (2023)

\bibitem{kumar2023generative}
Kumar, A., Bhunia, A.K., Narayan, S., Cholakkal, H., Anwer, R.M., Khan, S., Yang, M.H., Khan, F.S.: Generative multiplane neural radiance for 3d-aware image generation. In: Proceedings of the IEEE/CVF International Conference on Computer Vision. pp. 7388--7398 (2023)

\bibitem{kumar2023cross}
Kumar, A., Bhunia, A.K., Narayan, S., Cholakkal, H., Anwer, R.M., Laaksonen, J., Khan, F.S.: Cross-modulated few-shot image generation for colorectal tissue classification. In: International Conference on Medical Image Computing and Computer-Assisted Intervention. pp. 128--137. Springer (2023)

\bibitem{kumar2021udbnet}
Kumar, A., Ghose, S., Chowdhury, P.N., Roy, P.P., Pal, U.: Udbnet: Unsupervised document binarization network via adversarial game. In: 2020 25th International Conference on Pattern Recognition (ICPR). pp. 7817--7824. IEEE (2021)

\bibitem{li2023blip}
Li, J., Li, D., Savarese, S., Hoi, S.: Blip-2: Bootstrapping language-image pre-training with frozen image encoders and large language models. arXiv preprint arXiv:2301.12597  (2023)

\bibitem{li2023gligen}
Li, Y., Liu, H., Wu, Q., Mu, F., Yang, J., Gao, J., Li, C., Lee, Y.J.: Gligen: Open-set grounded text-to-image generation. In: Proceedings of the IEEE/CVF Conference on Computer Vision and Pattern Recognition. pp. 22511--22521 (2023)

\bibitem{lian2023llm}
Lian, L., Li, B., Yala, A., Darrell, T.: Llm-grounded diffusion: Enhancing prompt understanding of text-to-image diffusion models with large language models. arXiv preprint arXiv:2305.13655  (2023)

\bibitem{lin2014microsoft}
Lin, T.Y., Maire, M., Belongie, S., Hays, J., Perona, P., Ramanan, D., Doll{\'a}r, P., Zitnick, C.L.: Microsoft coco: Common objects in context. In: Computer Vision--ECCV 2014: 13th European Conference, Zurich, Switzerland, September 6-12, 2014, Proceedings, Part V 13. pp. 740--755. Springer (2014)

\bibitem{liu2023visual}
Liu, H., Li, C., Wu, Q., Lee, Y.J.: Visual instruction tuning. arXiv preprint arXiv:2304.08485  (2023)

\bibitem{lu2022unified}
Lu, J., Clark, C., Zellers, R., Mottaghi, R., Kembhavi, A.: Unified-io: A unified model for vision, language, and multi-modal tasks. arXiv preprint arXiv:2206.08916  (2022)

\bibitem{nichol2021glide}
Nichol, A., Dhariwal, P., Ramesh, A., Shyam, P., Mishkin, P., McGrew, B., Sutskever, I., Chen, M.: Glide: Towards photorealistic image generation and editing with text-guided diffusion models. arXiv preprint arXiv:2112.10741  (2021)

\bibitem{openai2023gpt}
OpenAI, R.: Gpt-4 technical report. arxiv 2303.08774. View in Article  (2023)

\bibitem{ouyang2022training}
Ouyang, L., Wu, J., Jiang, X., Almeida, D., Wainwright, C., Mishkin, P., Zhang, C., Agarwal, S., Slama, K., Ray, A., et~al.: Training language models to follow instructions with human feedback. Advances in Neural Information Processing Systems  \textbf{35},  27730--27744 (2022)

\bibitem{radford2021learning}
Radford, A., Kim, J.W., Hallacy, C., Ramesh, A., Goh, G., Agarwal, S., Sastry, G., Askell, A., Mishkin, P., Clark, J., et~al.: Learning transferable visual models from natural language supervision. In: International conference on machine learning. pp. 8748--8763. PMLR (2021)

\bibitem{ramesh2022hierarchical}
Ramesh, A., Dhariwal, P., Nichol, A., Chu, C., Chen, M.: Hierarchical text-conditional image generation with clip latents, 2022. URL https://arxiv. org/abs/2204.06125  \textbf{7} (2022)

\bibitem{rombach2022high}
Rombach, R., Blattmann, A., Lorenz, D., Esser, P., Ommer, B.: High-resolution image synthesis with latent diffusion models. In: Proceedings of the IEEE/CVF conference on computer vision and pattern recognition. pp. 10684--10695 (2022)

\bibitem{saharia2022photorealistic}
Saharia, C., Chan, W., Saxena, S., Li, L., Whang, J., Denton, E.L., Ghasemipour, K., Gontijo~Lopes, R., Karagol~Ayan, B., Salimans, T., et~al.: Photorealistic text-to-image diffusion models with deep language understanding. Advances in Neural Information Processing Systems  \textbf{35},  36479--36494 (2022)

\bibitem{shamshad2023clip2protect}
Shamshad, F., Naseer, M., Nandakumar, K.: Clip2protect: Protecting facial privacy using text-guided makeup via adversarial latent search. In: Proceedings of the IEEE/CVF Conference on Computer Vision and Pattern Recognition. pp. 20595--20605 (2023)

\bibitem{sharma2018conceptual}
Sharma, P., Ding, N., Goodman, S., Soricut, R.: Conceptual captions: A cleaned, hypernymed, image alt-text dataset for automatic image captioning. In: Proceedings of the 56th Annual Meeting of the Association for Computational Linguistics (Volume 1: Long Papers). pp. 2556--2565 (2018)

\bibitem{stiennon2020learning}
Stiennon, N., Ouyang, L., Wu, J., Ziegler, D., Lowe, R., Voss, C., Radford, A., Amodei, D., Christiano, P.F.: Learning to summarize with human feedback. Advances in Neural Information Processing Systems  \textbf{33},  3008--3021 (2020)

\bibitem{su2022language}
Su, Y., Lan, T., Liu, Y., Liu, F., Yogatama, D., Wang, Y., Kong, L., Collier, N.: Language models can see: Plugging visual controls in text generation. arXiv preprint arXiv:2205.02655  (2022)

\bibitem{taori2023stanford}
Taori, R., Gulrajani, I., Zhang, T., Dubois, Y., Li, X., Guestrin, C., Liang, P., Hashimoto, T.B.: Stanford alpaca: An instruction-following llama model (2023)

\bibitem{touvron2023llama}
Touvron, H., Lavril, T., Izacard, G., Martinet, X., Lachaux, M.A., Lacroix, T., Rozi{\`e}re, B., Goyal, N., Hambro, E., Azhar, F., et~al.: Llama: Open and efficient foundation language models. arXiv preprint arXiv:2302.13971  (2023)

\bibitem{tsimpoukelli2021multimodal}
Tsimpoukelli, M., Menick, J.L., Cabi, S., Eslami, S., Vinyals, O., Hill, F.: Multimodal few-shot learning with frozen language models. Advances in Neural Information Processing Systems  \textbf{34},  200--212 (2021)

\bibitem{vaswani2017attention}
Vaswani, A., Shazeer, N., Parmar, N., Uszkoreit, J., Jones, L., Gomez, A.N., Kaiser, {\L}., Polosukhin, I.: Attention is all you need. Advances in neural information processing systems  \textbf{30} (2017)

\bibitem{wu2023next}
Wu, S., Fei, H., Qu, L., Ji, W., Chua, T.S.: Next-gpt: Any-to-any multimodal llm. arXiv preprint arXiv:2309.05519  (2023)

\bibitem{yang2023reco}
Yang, Z., Wang, J., Gan, Z., Li, L., Lin, K., Wu, C., Duan, N., Liu, Z., Liu, C., Zeng, M., et~al.: Reco: Region-controlled text-to-image generation. In: Proceedings of the IEEE/CVF Conference on Computer Vision and Pattern Recognition. pp. 14246--14255 (2023)

\bibitem{yasunaga2023retrieval}
Yasunaga, M., Aghajanyan, A., Shi, W., James, R., Leskovec, J., Liang, P., Lewis, M., Zettlemoyer, L., Yih, W.t.: Retrieval-augmented multimodal language modeling  (2023)

\bibitem{you2023cobit}
You, H., Guo, M., Wang, Z., Chang, K.W., Baldridge, J., Yu, J.: Cobit: A contrastive bi-directional image-text generation model. arXiv preprint arXiv:2303.13455  (2023)

\bibitem{zhang2023adding}
Zhang, L., Rao, A., Agrawala, M.: Adding conditional control to text-to-image diffusion models. In: Proceedings of the IEEE/CVF International Conference on Computer Vision. pp. 3836--3847 (2023)

\bibitem{zhang2018unreasonable}
Zhang, R., Isola, P., Efros, A.A., Shechtman, E., Wang, O.: The unreasonable effectiveness of deep features as a perceptual metric. In: Proceedings of the IEEE conference on computer vision and pattern recognition. pp. 586--595 (2018)

\bibitem{zhang2022opt}
Zhang, S., Roller, S., Goyal, N., Artetxe, M., Chen, M., Chen, S., Dewan, C., Diab, M., Li, X., Lin, X.V., et~al.: Opt: Open pre-trained transformer language models. arXiv preprint arXiv:2205.01068  (2022)

\bibitem{zhang2023llavar}
Zhang, Y., Zhang, R., Gu, J., Zhou, Y., Lipka, N., Yang, D., Sun, T.: Llavar: Enhanced visual instruction tuning for text-rich image understanding. arXiv preprint arXiv:2306.17107  (2023)

\bibitem{zhou2022learning}
Zhou, K., Yang, J., Loy, C.C., Liu, Z.: Learning to prompt for vision-language models. International Journal of Computer Vision  \textbf{130}(9),  2337--2348 (2022)

\bibitem{zhu2023minigpt}
Zhu, D., Chen, J., Shen, X., Li, X., Elhoseiny, M.: Minigpt-4: Enhancing vision-language understanding with advanced large language models. arXiv preprint arXiv:2304.10592  (2023)

\end{thebibliography}
\end{document}